# A Multispectral Automated Transfer Technique (MATT) for machine-driven image labeling utilizing the Segment Anything Model (SAM)

James E. Gallagher, Aryav Gogia, Edward J. Oughton, *Member, IEEE*

Corresponding author: Dr. Edward J. Oughton (e-mail: eoughton@gmu.edu).

*Abstract*— Segment Anything Model (SAM) is drastically accelerating the speed and accuracy of automatically segmenting and labeling large Red-Green-Blue (RGB) imagery datasets. However, SAM is unable to segment and label images outside of the visible light spectrum, for example, for multispectral or hyperspectral imagery. Therefore, this paper outlines a method we call the Multispectral Automated Transfer Technique (MATT). By transposing SAM segmentation masks from RGB images we can automatically segment and label multispectral imagery with high precision and efficiency. For example, the results demonstrate that segmenting and labeling a 2,400-image dataset utilizing MATT achieves a time reduction of 87.8% in developing a trained model, reducing roughly 20 hours of manual labeling, to only 2.4 hours. This efficiency gain is associated with only a 6.7% decrease in overall mean average precision (mAP) when training multispectral models via MATT, compared to a manually labeled dataset. We consider this an acceptable level of precision loss when considering the time saved during training, especially for rapidly prototyping experimental modeling methods. This research greatly contributes to the study of multispectral object detection by providing a novel and open-source method to rapidly segment, label, and train multispectral object detection models with minimal human interaction. Future research needs to focus on applying these methods to (i) space-based multispectral, and (ii) drone-based hyperspectral imagery.

## I. INTRODUCTION

Segmentation is a processing step used to partition an image into different homogeneous regions or clusters [1]. These homogenous pixel clusters, typically object classes, provide meaningful results when applying computer vision to understand images or video. Such techniques are critical in a range of consumer industrial and military applications [2]. However, extensive labor is traditionally required, as segmentation involves manually outlining each unique object class via a labeling tool (essentially drawing a rectangular box around a set of pixels and providing a text classification). Human-driven labeling is one of the most significant resource barriers to developing computer vision models [3], especially in multispectral bands. Labeling takes time, and the accuracy of the segmentation mask will also vary due to human error [4]. The Segment Anything Model (SAM) has opened a new frontier of possibilities for accurate and consistent image segmentation performance of Red-Green-Blue (RGB) images [5]. With SAM, what may have taken months of work to segment a dataset can be rapidly reduced into hours or days.

Consequently, the speed and accuracy of automatically segmenting large RGB datasets has been drastically accelerated. SAM was trained on a dataset consisting of 11 million images and 1.1 billion masks, with powerful zero-shot performance on a range of segmentation tasks [6]. However, SAM's primary limitation is that this tool can currently only be applied to segmenting object classes in visible RGB images, without any comparable tool for multispectral imagery. On the left-hand diagram of Fig. 1B we illustrate SAM's ability to segment cars in RGB, but lack thereof when applied to a comparable Long-Wave Infrared (LWIR) image. Conducting image segmentation on multispectral images, particularly for LWIR, is of critical importance to the scientific community because we can use different frequency bands to enhance object detection performance. Therefore, building automatic segmentation techniques for LWIR object detection models is an attractive area of current research [7]. For example, unlike RGB sensors which perform optimally during daytime, LWIR sensors work in low visibility conditions, highlighting the need to advance LWIR object detection research [8], [9]. Indeed, such sensors are increasingly used in a range of applications such as autonomous driving, military surveillance, search and rescue operations, conservation efforts, and infrastructure assessments [10], [11], [12], [13]. Without new scientific approaches to machine-driven labeling, the tedious task of manually segmenting images will prevent more effective multispectral object detection models from being deployed *at scale*.

Given this limitation, we fill a critical capability gap within the field of multispectral object detection and image segmentation by creating a novel method called the Multispectral Automated Transfer Technique (MATT). This approach utilizes an RGB segmentation mask from SAM and transposes the edge outline onto paired multispectral imagery [14]. Fig. 2 illustrates the proposed solution using drone-based imagery. As seen in Fig. 3, applying MATT to large multispectral datasets can rapidly create consistent label and segmentation performance to train object detection models, motivating the need for basic research that scientifically quantifies the impacts of different key variables on performance. The primary benefit of MATT is its ability to retrain new multispectral models in near-real time. This enables the end-user to rapidly create and deploy object detection models trained on relevant data. MATT not only allows for rapid model training, but also provides the ability to generate enormous multispectral datasets at machine speed. Although MATT can be used for both ground and air-based imagery, this research will evaluate MATT using drone-based imagery, which is inherently more challenging due to a variety of variables, such as elevation and illumination levels [15].

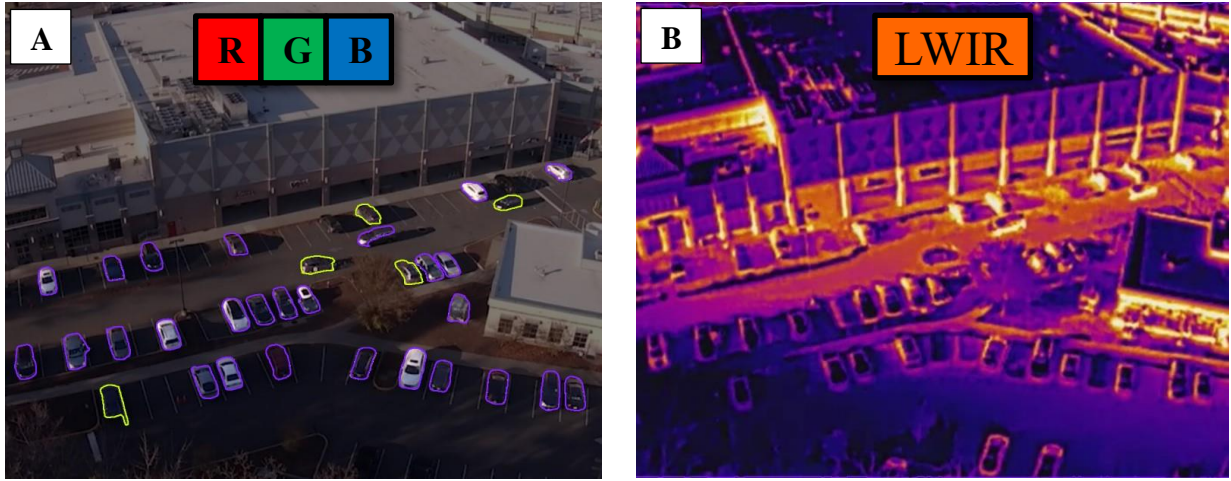

**Fig. 1.** SAM was applied to both RGB and LWIR images. As seen in Fig. 1A, SAM adequately segments vehicles. However, no vehicles were segmented by SAM in the LWIR image.

As a drone's elevation increases, sensor resolution deteriorates, with certain sensor resolutions fading faster than others [16]. Additionally, illumination levels and time-of-day play an important role in segmentation [17]. Shadows cast by an object class during Post-Sunrise or Pre-Sunset hours will affect segmentation performance, as edge detection becomes increasingly less defined due to shadows, thus deteriorating object detection performance [18], [19].

Training an object detection model on multispectral imagery typically takes a significant amount of time to complete individual necessary tasks, such as image extraction, image processing, segmentation, labeling, and model training. However, MATT not only automatically segments and labels images from any spectral band, but it also has built-in features to decrease the time required to process, label, train, and deploy the given model. In light of this, we will investigate the following two scientific research questions:

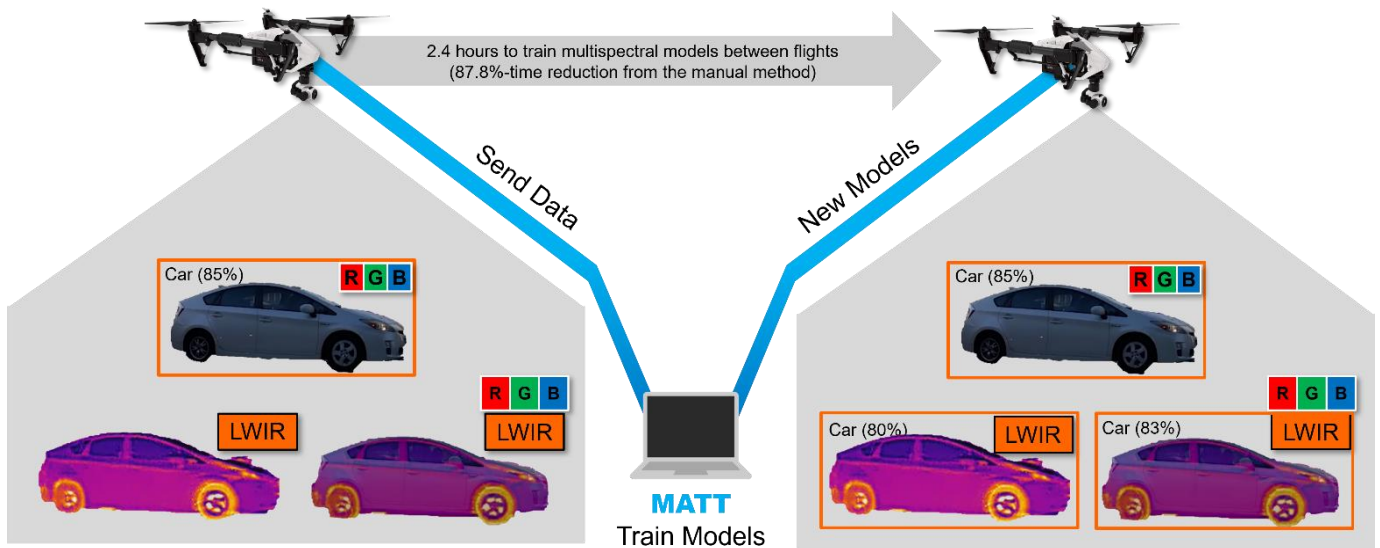

**Fig. 2**. Proposed Solution.

TABLE I
Abbreviations Used in this Research

| Abbreviation | Definition |
| --- | --- |
| AVIRIS | Airborne Visible InfraRed Imaging Spectrometer |
| CNN | Convolutional Neural Network |
| FAA | Federal Aviation Administration |
| FLIR | Forward-Looking Infrared |
| FOV | Field of View |
| FPS | Frames Per Second |
| GPU | Graphics Processing Unit |
| HSI | Hyperspectral Imagery |
| LWIR | Long-Wave Infrared |
| LWIR MATT | Long-Wave Infrared Multispectral Automated Transfer Technique |
| mAP | Mean Average Precision |
| MATT | Multispectral Automated Transfer Technique |
| MSI | Multispectral Imagery |
| NASA | National Aeronautics and Space Administration |
| RGB | Red-Green-Blue |
| RGB MATT | Red-Green-Blue Multispectral Automated Transfer Technique |
| RGB-LWIR | Red-Green-Blue-Long-Wave Infrared |
| RGB-LWIR MATT | Red-Green-Blue-Long-Wave Infrared Multispectral Automated Transfer Technique |
| SAM | Segment Anything |
| SODA | Segmenting Objects in Day And night |
| YOLO | You Only Look Once |

1. How effective is MATT compared to manual labeling approaches?

2. What impact does elevation, time-of-day and sensor type have on models trained either via a manually labeled dataset or MATT?

This basic research provides scientific inight for how an automated-labeling approach can support a wide range of object detection use cases in consumer, industrial, and military applications. For example, if multispectral data is collected on animal specific object classes for conservation purposes, scientists could more efficiently create a model in the field and deploy it with high accuracy [20], [21]. In military applications, there is a growing demand for the capability to rapidly train and deploy an object detection model based on limited multispectral data of newly developed adversarial equipment entering the battle space [22], [23]. Equally, for humanitarian missions, mine detection in conflict or post-conflict regions is an important research area [24], [25], given the cost and risk of manual approaches [26].

The benefit of using MATT is that the approach makes it easy to train and deploy, especially for real-time implementation with data. This research will provide key scientific contributions by quantifying drone-based object detection model performance when presented with manually labeled datasets versus automatically segmented datasets using three sensor types.

In the following Section, a literature review is undertaken, with the method then presented in Section III, before returning to results in Section IV. Finally, Section V will discuss the findings, while Section VI will provide concluding remarks and the way ahead.

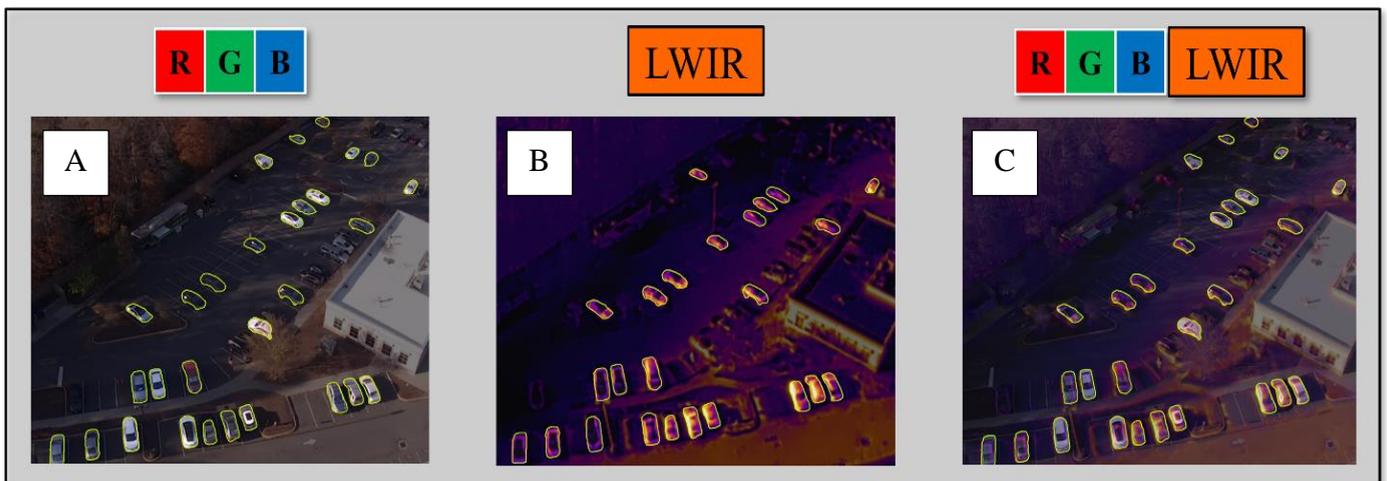

**Fig. 3.** Applying MATT to multispectral images produces consistent results.

## II. LITERATURE REVIEW

Since its release in 2023, SAM has been rapidly evolving, with the original arXiv paper cited over 1,600 times in 2023 alone. The method is being heavily utilized in experiments across a variety of different fields that are using object detection. However, despite the broad SAM literature, there is very little research that discusses applying SAM to drone-based or satellite-based Multispectral Imagery (MSI) and Hyperspectral Imagery (HSI). One research study has applied SAM to improve the segmentation of spectral clusters from HSI data derived from NASA's Airborne Visible/Infrared Imaging Spectrometer (AVIRIS) [27], [28]. Another study creates a semi-supervised method to label and classify HSI data to segment and monitor land use with high accuracy [29]. Yet, there is also minimal literature on drone-based LWIR object detection, although there is active research trying to creating methods to decrease long-distance LWIR resolution loss [30], [31].

In RGB drone-based applications, SAM is being utilized for measuring crop yields for agricultural purposes, and conducting tree-segmentation to measure forest density and health [32], [33], [34], [35], [36]. These methods are also being applied to understand the built environment, including infrastructure health [37], from railroad integrity, to bridge structures, and detecting road cracks [38], [39], [40], [41]. In the natural sciences, SAM is being applied to sonar imagery to enhance object detection in planetary observation [42].

The importance of multispectral imagery for object detection is well recognized. Firstly, RGB sensors are limited in their ability to work in low-visibility environments [43], [44]. Thus, integrating other spectral bands with RGB, like LWIR, enhances overall object detection performance by improving machine depth perception [45]. Indeed, LWIR sensors are excellent at identifying object classes in complex environments where visibility is limited [46]. According to recent research using an RGB-LWIR fused dataset for object detection applications outperforms an RGB dataset by 1.2% [47]. Furthermore, the RGB-LWIR fused dataset is highly resilient to changes in visibility and performs consistently during all hours of the day [48].

LWIR sensors are also heavily utilized in autonomous driving because of their ability to see in complex low visibility environments [49], [50] [51], [52], [53]. However, a fundamental issue in applying this sensor type in object detection use cases can be the low-resolution present in LWIR images and video footage. LWIR resolution deteriorates exponentially as the distance to the object class increases [54]. Therefore, fusing RGB images helps to minimize edge deterioration over distance.

The literature provides an abundance of research studies and use-cases for conducting both image segmentation and semantic segmentation on RGB-LWIR images, as well as the value provided in fusing these two image-types for object detection applications [55], [56]. Conventional semantic segmentation and image segmentation exploit the three-channel RGB image spectrum. However, as previously mentioned, RGB performs poorly during periods of limited visibility [57]. Given the nature of thermal sensors to perform in all visibility conditions, incorporating a paired LWIR image to an RGB image allows for overall greater segmentation accuracy in complex visibility conditions [47], [58].

The most common application for RGB-LWIR semantic segmentation is also in autonomous vehicles [59]. Extensive research has been conducted in how to best implement RGB-LWIR semantic segmentation to increase mean average precision (mAP) and intersection of union (IoU) results. One study created an architecture called an edge-conditioned convolutional neural network (EC-CNN), which is trained to optimize LWIR segmentation [60]. A recent study created an end-to-end deep neural network, called FuseSeg, where RGB-LWIR paired images are used to increase semantic segmentation of urban scenes [51]. The research also produced a dataset, called Segmenting Objects in Day And night (SODA), to promote further research in thermal semantic segmentation.

Research has been conducted in up-sampling low resolution images to increase segmentation accuracy [61]. Similarly, one study creates an RGB to infrared fusion module, named Attention Fusion Network (AFNet), to combine complementary characteristics from both image types [62].

Additionally, a recently published study produced the Multi-Expert Fusion Network (MEFNet), which selects the most important features from both RGB and LWIR images to create the highest quality segmentation [63]. Research has also been carried out to use RGB image segmentation to help fuse gaps produced in LWIR segmentation caused by transparent objects such as glass [64]. This allows the completion of LWIR segmentation masks when conducting thermal object detection in environments that have translucent objects. Similar concepts have also been presented where an adaptive RGB-LWIR fusion method is used to optimize segmentation results based on illumination [65].

The literature also illustrates that paired RGB-LWIR images increase semantic segmentation for an array of industrial applications. According to one study, applying edge detection techniques, such as sobel, canny and otsu, to extract key edges from RGB and LWIR pair images and overlaying the most critical portions can be used to measure the efficiency of industrial related activities [66]. Research has also been carried out on LWIR sensors to conduct semantic segmentation and edge extraction of heat-sources of interest to create an electrical thermal image segmentation dataset to determine infrastructure health [67].

SAM is also rapidly growing in the field of remote sensing, where this tool is being combined with a myriad of python-based geospatial packages to provide a straight-forward interface for users [68]. Additionally, data-pipelines are being built that leverage SAM to rapidly segment large quantities of remotely sensed data [69]. When analyzing the U.S.

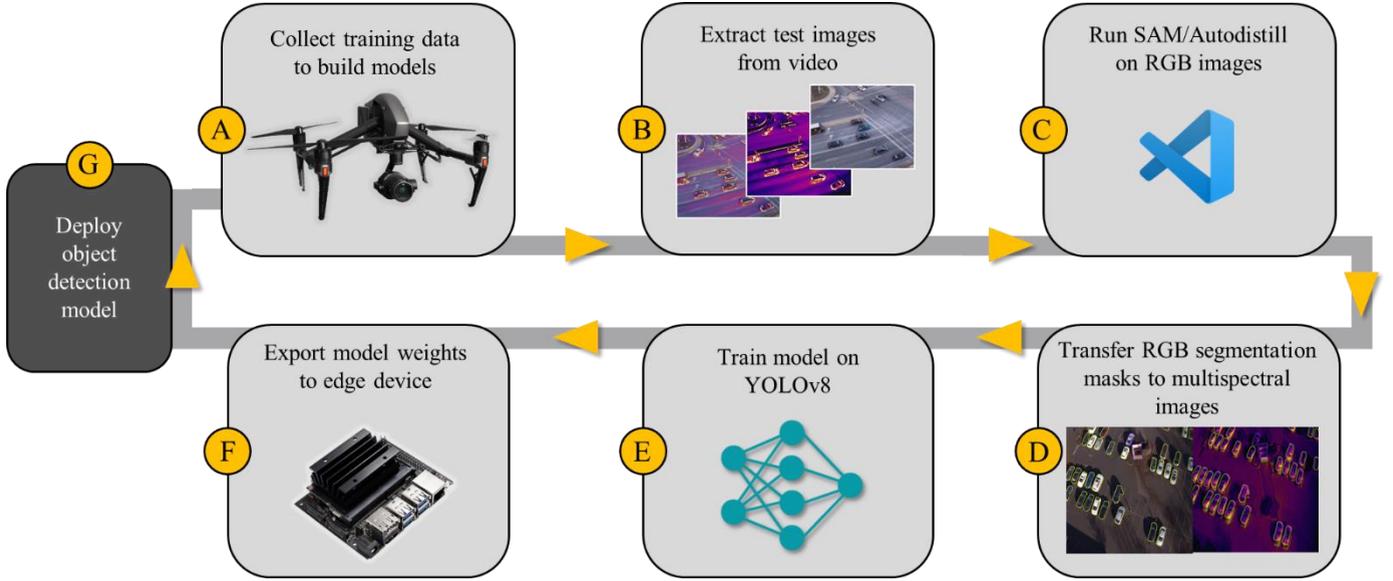

**Fig. 4.** Workflow for creating multispectral object detection models with MATT.

Department of Agriculture's Cropland Data Layer, SAM was used to better segment land-use and crop type being grown across the United States with high precision [70]. When conducting change detection in high-resolution remotely sensed images, SAM is being harnessed to more accurately and rapidly segment pixel-level changes to increase precision in change detection results [71]. For mapping applications, such as building segmentation and urban area segmentation, research is being conducted to combine SAM with remotely sensed satellite data [72], [73], [74]. There are also attempts on creating solutions to apply SAM on remotely sensed data with lower spatial resolutions, achieving relatively high segmentation success [75].

Considerable activity has also been taking place in utilizing SAM in the medical field to segment human organs and diseases in medical images [76], [77], [78], [79], [80]. The findings from the research conclude that SAM is superior at segmenting and identifying well-defined objects, such as organs, but sometimes has difficulty in identifying objects with unclear borders, such as tumors and cancer. New medical models that wield SAM, such as MedSAM, are promising improved results by outperforming existing state-of-the-art medical object detection models [81]. However, medical images captured in a controlled environment. There are similarities here with the way drone imagery is collected, given a range of other factors affect model performance (visibility, elevation, etc.).

In summary, there exists a shortage of openly available multispectral datasets for semantic segmentation [82], [83], driven by the labor-intensity of manual labeling. Due to this literature gap, more studies on RGB-LWIR semantic segmentation are required [84], motivating this research.

## II. METHOD

This section describes the methods and workflow for creating MATT, which will hitherto be referred to as the 'automated' method and compared to a 'manual' labeling method. The automated method will utilize the SAM generated RGB segmentation mask and overlay this onto paired multispectral imagery. The manual method is the traditional practice of a human using a labeling tool to outline the object class with a bounding box followed by applying a label to the bounding box. The bounding box and labeling process would be repeated for multiple object classes within an image. Fig. 4 outlines the general approach of this research, while Fig. 5 visualizes the workflow for both manual and automated methods. The first step in the research is to design a system workflow that will optimize the training of multispectral object detection models. The steps outlined in Fig. 4 will describe the key stages of MATT. Steps A and F are hardware-centric, while Steps B through E are code-centric for the automated method.

In step A, data collection will be conducted from an air-based platform. For this research, a DJI Inspire 2 multirotor drone carrying an RGB and LWIR sensor package will be used to collect multispectral data. The RGB camera selected is the RunCam 5 Orange. This camera is lightweight and designed for drone applications. The RunCam costs $110 USD and has a resolution of 1280 x 1024 pixels with a Field of View (FOV) of 145°. The RunCam capture settings, which include shutter speed, color style, saturation, exposure, contrast, sharpness, and white balance, will be set to the default values.

TABLE II
Variables that can be modified in MATT

| Variable | Default |
| --- | --- |
| Epochs | 200 |
| Object Class Ontology | "cars", "trucks" |
| Image Processing Steps | Turned off |
| Frame extraction rate (F-stride) | 100 |

The LWIR sensor that will be used is the Forward-Looking Infrared (FLIR) Vue Pro R. This is a radiometric sensor designed for drone applications that costs $2,900 USD. The FLIR Vue Pro R has a resolution of 640 x 512 pixels (exactly half the resolution of the RunCam RGB sensor) and a FOV of 45° with a lens diameter of 6.8mm. The 30 Hz variant of the FLIR Vue Pro R will be used. The spectral band that the FLIR Vue Pro R can capture is 7.5 - 13.5 µm with an operating temperature range of -20° C (-4° F) to 50° C (122° F) [85]. Default sensor settings will be kept on the FLIR Vue Pro. Lastly, the fusion thermal palette will be selected because of its properties that make it optimal to fuse with RGB. Resolution for RGB footage will be down-sampled and fused with LWIR footage in Adobe Premiere Pro, using a 50% fusion of both image types. Both sensors will also be coaligned to minimize parallax [86]. Video footage from both RGB and LWIR sensors will record at 30 Frames Per Second (FPS).

Table II outlines key parameters in MATT that can be modified. These parameters can be added or removed given desired object detection requirements. After collecting data, step B in Fig. 4 is the first stage in the automated method. During this step, frame extraction from post-flight drone footage is conducted. Footage is transferred from the sensors to the processing unit via removable micro-SD cards. In MATT, frame stride, also known as F-stride, can be set to control the rate of images to be extracted. Higher F-stride equates to fewer images extracted from the footage. For example, both the RGB and LWIR sensors record at 30 FPS. If the F-stride is set to a value of 30, one frame would be extracted every second. A lower F-stride results in more images and a larger training dataset. However, more images will also increase the time it will take to segment and label a dataset and train a model. The F-stride value determines the equilibrium between training speed and model accuracy. An F-stride of 100 is the default setting for the automated method (so a frame extraction every 3.3 seconds at 30 FPS).

During step C, RGB images will be processed through SAM for segmentation and through Autodistill for labeling. Autodistill is a package developed by Roboflow that provides automatic labeling using ontology, minimizing the need for complex code [87]. SAM only segments clusters into segmentation masks, whereas Autodistill provides labels to the segmentation masks. The caption ontology command in MATT is where the ontology can be modified to create a custom labeled dataset of the object class of interest. For this research, the ontology of "car" and "truck" will be used.

Prior to segmentation, image processing can be conducted within MATT. The automated method has six built-in image processing and edge enhancement techniques that can be utilized to grow the training dataset prior to image segmentation. The MATT image processing filters are commented out in the default package. The image processing filters available in the automated method include flipping, blurring, flipping and blurring, SobelXY, Distribution of Gaussian, and Gaussian Threshold. Image processing not only increases the size of the training dataset, but also allows the model to train on more complex images, thereby increasing performance. The blurring, as well as the blurring and flipping, image augmentation techniques are especially beneficial because of blurry footage and images caused by drone movement and vibrations due to oscillatory motions [37].

In step D, the newly created RGB segmentation masks and labels in the form of .txt files will then be transferred to their respective LWIR and RGB-LWIR folders. The YOLO .txt label file produced from RGB segmentation is associated to the matching LWIR and RGB-LWIR paired image through the commonly shared file name. The segmentation mask aligns seamlessly with the multispectral images providing that the footage from all three image types are extracted with the same F-stride. This process can be summarized in eq. (1), where *SAM* represents the segmentation masks generated from RGB images, while $YOLOv^8$ denotes the CNN. $TF(SAM(^XRGB),^XMS)$ represents the transfer of segmentation masks generated by SAM from the RGB Images ($^XRGB$) top the multispectral images ($^XMS$). Lastly, $^Tm$ represents the time required to manually verify the segmentation masks.

$$y_{ms} = YOLOv^8(TF(SAM(x_{RGB}), x_{MS})) + T_m \quad (1)$$

Equation (2) demonstrates the manual method for conducting the segmentation and labeling of images, where $L(^XMS)$ denotes the manual labeling process applied to multispectral images ($^XMS$). Lastly, $^Ymanual$ is the output detection results using the manual method.

$$y_{manual} = YOLOv^8(L(x_{MS})) \quad (2)$$

Step E is the final step in the automated process. Here, the newly labeled multispectral dataset can be passed through the neural network to begin training. YOLOv8 will be selected as the pre-trained Convolutional Neural Network (CNN). YOLOv8 is selected for MATT because the model is the fastest and most accurate open-source CNN at the time of this research [88]. Although performance gains between YOLOv8 and YOLOv7 are minuscule, YOLOv8 is much easier to integrate with various software and packages. There are five models in YOLOv8, ranging from nano to extra-large. The small model, YOLOv8s, will be used as the default model in the automated method due to its reasonable speed and relative accuracy [88]. YOLOv8s will also be the CNN used in this research to measure model performance. The default number of training epochs in MATT is set to 200. Following the completion of model training, the weights will then be exported to the file path destination established in the automated method. Finally, Steps F and G involve deploying the weights to the edge device. To save time, MATT should run on a Graphics Processing Unit (GPU) edge device. This will reduce the need to export model weights from a desktop computer to an edge device for deployment.

For this research, model training and testing will each utilize their own distinct datasets, preventing the artificial inflation of mAP by ensuring models are evaluated with never-before-seen images. The training dataset will consist of 2,400 images. Images will be collected during various hours of the day at different elevations to ensure image heterogeneity between high and low illumination periods. The two object classes of interest for this research are cars and trucks. A total of 100 original images are collected for each object class, equating to 200 original images per sensor type. Three image processing techniques will then be used to grow the dataset to 800 images per sensor type. The three image processing techniques used are (i) blurring, (ii) flipping, and (iii) blurring and flipping. For the research experiment, two sets of object detection models will be created. The first set of models will be trained with images that are manually labeled with bounding boxes. The second set of models will be trained using image segmentation and labels generated by the automated method.

After the three automated and manually labeled models are trained, a separate testing dataset of 1,200 images will be used to measure model performance using mAP as the primary benchmark. Fig. 5 visualizes the experiment and variables. The testing dataset will consist of images collected at fixed elevations and various periods of the day to measure detection performance against elevation and illumination variables. Five images will be collected at each elevation at a different time-of-day, totaling 1,200 images, allowing for sufficient confidence intervals to be calculated for final results. Images for the testing dataset will be collected at 15 m (50 ft), 30 m (100 ft), 45 m (150 ft), 61 m (200 ft), 76 m (250 ft), 91 m (300 ft), 106 m (350 ft), and 121 m (400 ft). Moreover, images will be collected at five different periods of the day. These periods are Pre-Sunrise (low-thermal cross-over, low illumination), Post-Sunrise (low-thermal cross-over, medium illumination), Noon (high-thermal cross-over, high illumination), Pre-Sunset (high-thermal cross-over, medium illumination) and Post-Sunset (medium-thermal cross-over, low illumination).

The last variable that will be measured is time taken to conduct manual labeling when compared to the automated method. To measure this, 100 images will be labeled manually with a common labeling tool, LabelImg. Time will be measured by how long it will take to label each image. The images selected will be drawn from the training dataset, and will vary in elevation, time-of-day, and sensor type. After the hundred images are labeled, the average time to label the 100 images will be multiplied by twenty-four to represent the average time it would take to label all 2,400 images in the training dataset. The time constraints for manual labeling can be broken down in eq. (3), where $T$ represents the total time to manually label images, n is the number of images, $t_i$ is the time to label each $i^{th}$ image.

$$T = \sum_{i=1}^{n} t_i \qquad (3)$$

Since the average time to manually label an image was 30 seconds, the time to approximate total manual labeling time for 2,400 images can be explained in eq. (4).

$$T = 20 \ hours = \sum_{i=1}^{2400} 30 \qquad (4)$$

This time is an estimate, as breaks and rests that would be required for a human conducting manual labeling are not accounted for in this equation. MATT will be measured using a python module called timeit to measure the execution time of the code. MATT will also be run on a CPU (Apple M1) and GPU (NVIDIA T4) to measure performance differences.

### III. RESULTS

When analyzing time-of-day performance, Fig. 6 visualizes model performance for each sensor type. Each model and sensor were tested against 40 test images for each time-of-day to generate mAP. Error bars were generated by using mean and standard error values. The lower limit of the error bar is calculated by subtracting the standard error from the mean, while the upper limit of the error bar is calculated by adding standard error to the mean. The overall mAP of models trained on manually labeled datasets was 61.9% (reaching 68.7% excluding nighttime data), while the overall mAP of models trained with the automated method (MATT) was 55.2% (reaching 62.4% excluding nighttime). Thus, the automated method helps to reduce labeling time by 87.8%, with only a 6.7% decrease in performance. Interestingly though, when analyzing Fig. 6 to observe individual sensor performance during different times-of-day, the automated method applied to the RGB sensor actually *outperformed* all other models during daytime periods. An overall average mAP of 81.7% was achieved when conducting object detection between 16 m and 121 m, while the RGB manual method achieved a mAP of only 75.5%.

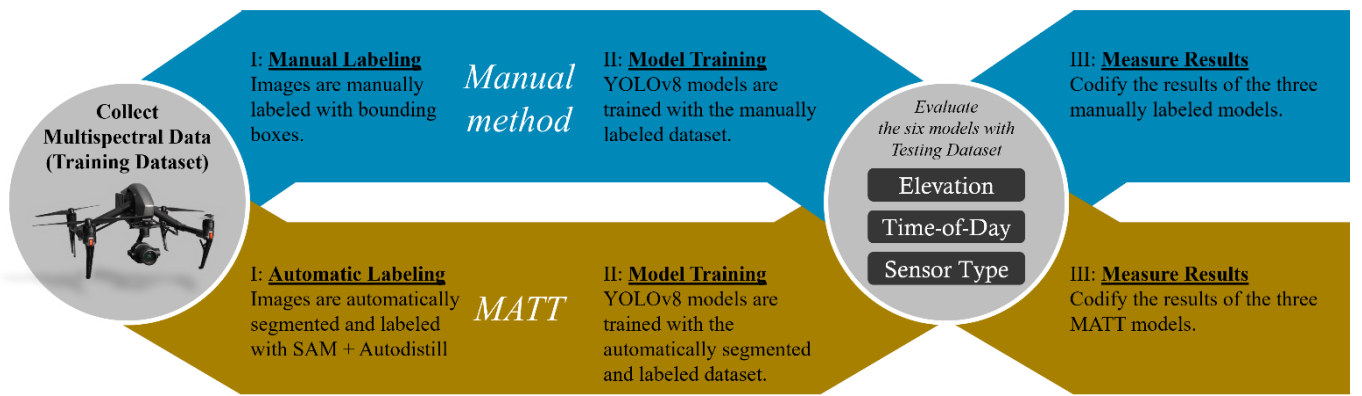

**Fig. 5**. The research approach given key uncertainty factors, including elevation, time-of-day, and sensor type.

When applying different labeling techniques to the LWIR sensor type, we find that manual models consistently outperformed the automated labeling approach. during all periods of the day. The LWIR manual method attained an average mAP of 59.3% while the LWIR automated method performed significantly less with an average mAP of 42.5%, accounting for a 16.9% decrease in overall performance when using the automated method for LWIR object detection. Among the three sensor-types the LWIR automated model suffered the largest performance drop when compared to the RGB and RGB-LWIR automated models.

The RGB-LWIR models performed slightly below the RGB models, with the RGB-LWIR manual model scoring a mAP of 64.9% and the automated method with a mAP of 56.3%, resulting in an 8.6% drop in performance using the automated method. When comparing the automated and manual models for LWIR, which performed with a similar performance gap during all times of the day, the fused RGB-LWIR method only had a slight performance difference during most periods of the day except for Post-Sunset periods. The RGB-LWIR model only suffered a 6.0% loss in performance if Post-Sunset performance data is excluded. During the Post-Sunset period, the RGB-LWIR automated model performed 18.9% poorer than the RGB-LWIR manual method.

When analyzing performance loss between the manual and automated approaches across different periods of the day, the Post-Sunset period suffered the highest loss in performance for automated models. The Post-Sunset period accounted for a decrease in 10.8% mAP when compared to manual model performance. The period of the day that had the lowest loss in automated performance was the Pre-Sunset period with only a 4.5% loss in mAP when compared to the manual approach. Pre-Sunrise, Post-Sunrise and Noon periods had a 6.9%, 6.6% and 4.8% loss in performance between the manual and automated approaches. Conversely, the sensor type and model that had the largest mAP improvement in relation to time-of-day was the RGB automated model during Pre-Sunset hours, outperforming the RGB manual approach by 9.3%. The RGB-LWIR automated model had the smallest performance gap during Pre-Sunset, falling only 5.3% behind the RGB-LWIR manual model.

Noon period also had the highest performance of all manual and automated methods. The highest mAP for the automated method was with the RGB sensor at Noon, achieving an average mAP of 84.3%. LWIR and RGB-LWIR also performed optimally during Noon, achieving a mAP of 49.7% and 61.5% respectively. Noon and Pre-Sunset periods were also the superior periods to deploy automated models, with these approaches only suffering a 4.8% (Noon) and 4.5% (Pre-Sunset) lose during these periods.

Pre-Sunrise and Post-Sunset periods were aggregated to analyze nighttime performance, while Post-Sunrise, Noon, and Pre-Sunset periods were combined to analyze daytime performance. For daytime performance, the elevation with the highest overall average performance was 47 m, achieving a mAP of 81.8%, with the second highest performance benchmark at 31 m with 77.7% mAP. The daytime elevation with the lowest mAP performance was 121 m with an average mAP of 48.8%, followed by 16 m with an average mAP of 50.6%. The highest mAP achieved in this research was by the automated method combined with RGB at 31 m, achieving a mAP of 93.9%. The next highest performing automated RGB instance was at 47 m at a 92.2% mAP.

The best performing manual RGB approach was at 31 m at 91.4% mAP. The most significant performance gap between the RGB manual and automated model was at 121 m, where the RGB automated model outperformed the RGB manual approach by 13.2%. The lowest mAP for both RGB manual and RGB automated was at 121 m with 50.3% and 63.5% mAP. Both the LWIR manual and automated approaches performed the worst during daytime, achieving an average mAP of 60.6% for manual and 45.2% for the automated approach. The highest performing daytime elevation for manual and automated LWIR was at 47 m with a mAP of 78.5% and 67.7%. The worst daytime elevation for both LWIR approaches was 16 m, with LWIR manual achieving 33.6% mAP and LWIR automated with 18.4% mAP. The largest daytime elevation model performance gap was at 31 m, where the LWIR manual method outperformed the automated approach by 27.8%.

The RGB-LWIR models were the second-highest performing during daytime, with the RGB-LWIR manual method scoring an average mAP of 66.9% and the automated RGB-LWIR approach achieving a mAP of 60.7%. Like LWIR, the elevation with the most optimal performance for both RGB-LWIR approaches was at 47 m with 85.0% for the RGB-LWIR manual model and 78.8% for the RGB-LWIR automated method. The largest performance difference between the RGB-LWIR manual and automated models was also at 16 m, with the RGB-LWIR manual model outperforming the automated method by 22.0%. Conversely, the smallest performance gap was at 109 m, where the manual model outperformed the RGB-LWIR automated method by only 2.1%.

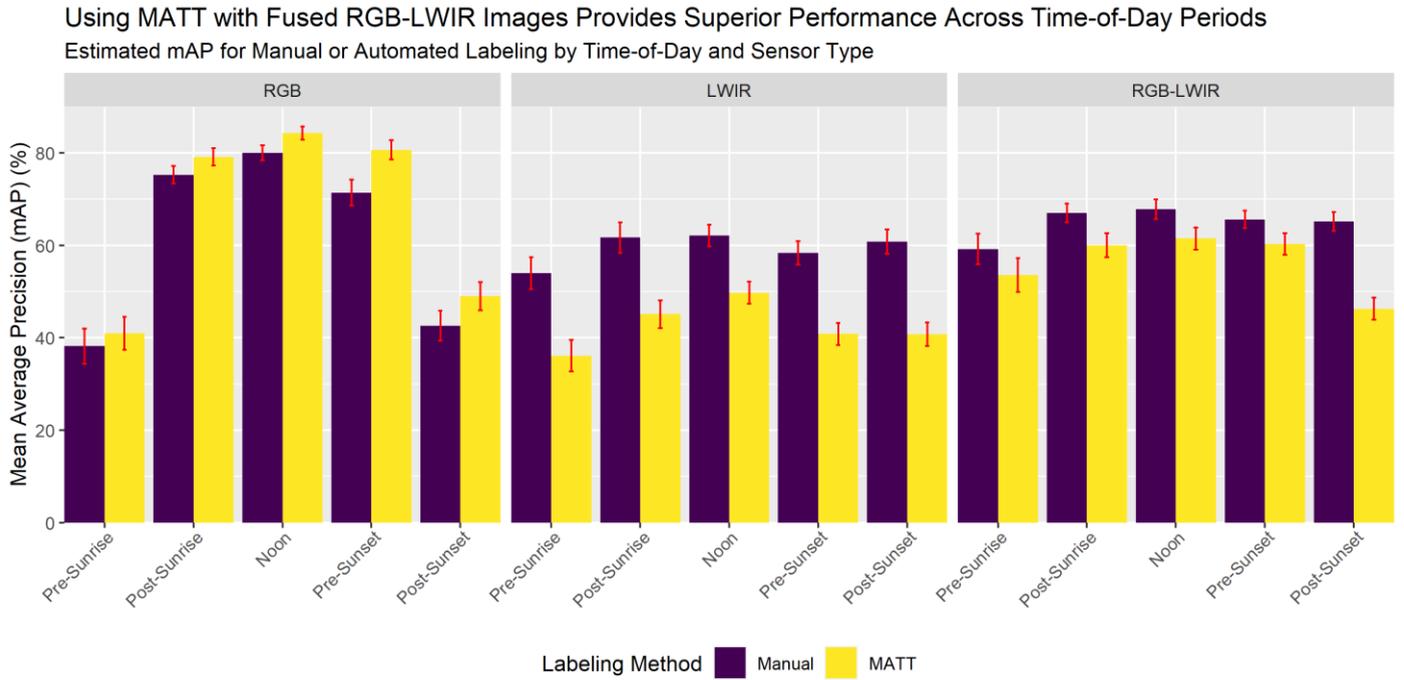

**Fig. 6.** Comparing model performance for manual versus MATT labeling methods by sensor type and time-of-day.

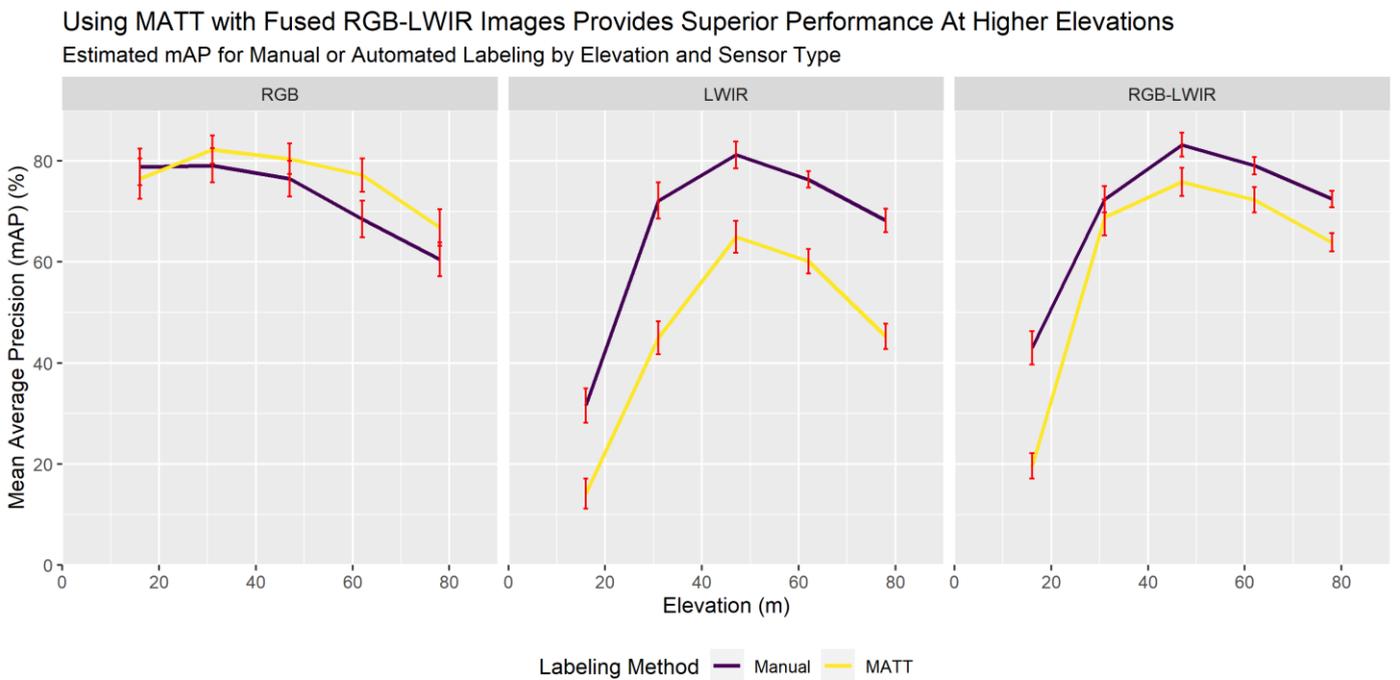

**Fig. 7.** Comparing model performance for manual versus MATT labeling methods by sensor type and elevation.

For nighttime performance, the elevation with the highest overall average performance was 47 m with an average mAP of 69.8%, followed by 62 m with an average mAP of 69.5%. The elevation with the worst overall nighttime mAP was 121 m with an average mAP of 26.5%, followed by 109 m with 34.4%. Interestingly, like daytime performance, the lowest elevation of 16 m did not yield the highest mAP, only achieving 34.0% at nighttime. The elevation with the highest nighttime average mAP for manual models was 47 m with 74.7%, followed by 62 m with 71.2%. The elevation with the greatest automated method performance was 62 m with 67.7%, followed by 47 m with 65.0%. The largest average nighttime performance gap between the manual and automated models was at 16 m. Furthermore, the average difference was 18.9% between manual and automated model performance. The worst elevation for both manual and automated model performance was 121 m, with models performing at 29.5 and 23.6%, respectively.

Although RGB would not be the sensor of choice for nighttime object detection, the automated RGB model outperformed the RGB manual method at every elevation except at 16 m, where the RGB manual approach outperformed the automated method by 10.1%. The greatest performance gap between manual and automated models at nighttime was at 94 m, where the automated RGB method outperformed the manual method by 12.3% mAP. Additionally, the highest performing RGB model at nighttime was the RGB manual approach at 16 m with a mAP of 64.2%. Overall, RGB performed the worst at 109 m and 121 m, where mAP ranged between 9.3% and 16.3% for both manual and automated models.

The LWIR manual model was the second-highest performing model at night, with an overall average mAP of 57.4%, while the average automated LWIR mAP was 38.4%. The best elevation for the LWIR manual method was 47 m with a score of 85.3%, while the highest performing nighttime elevation for automated LWIR was 62 m with a mAP of 65.7%. The greatest performance difference between the manual LWIR and automated LWIR model was at 31 m, with the manual approach outperforming the automated LWIR approach by 26.2%. The elevation that resulted in the lowest nighttime performance was 16 m, with manual LWIR performing at 28.6% and automated LWIR at 7.8%.

The RGB-LWIR manual model had the highest mAP at nighttime with an average mAP of 62.2%. The automated RGB-LWIR method had an average nighttime performance of 49.9% mAP. Furthermore, the best performing nighttime elevation for automated RGB-LWIR was at 62 m with a mAP of 83.8%, while the greatest nighttime performance for the RGB-LWIR manual approach was also at 62 m achieving a mAP of 75.2%. Like LWIR, the worst nighttime elevation for both RGB LWIR manual and automated approaches was 16 m, with a mAP of 37.3% and 11.8%. This elevation also had the largest nighttime performance gap between the RGB-LWIR manual and automated models, resulting in a 25.6% difference in performance.

Lastly, Table III compares time used to create models with the manual and automated methods. Labeling the training dataset manually took approximately 30 seconds per image, or 20.0 hours. This is the total time it would take to label the training dataset if it were conducted non-stop by a single human. Using a CPU, it took MATT 1.1 hours (66 minutes and four seconds) to segment and label the entire training dataset. When using a GPU, it only took .22 hours (13.5 minutes) for MATT to segment and label the dataset. Additionally, 2.2 hours (10 seconds per image) was required for a human to manually verify and conduct minor label corrections for the MATT dataset, resulting in a total of 2.4 hours for the automated method.

## IV. DISCUSSION

This study first undertook an assessment to quantify the impacts of using SAM to automate the labeling of multispectral images. Secondly, this research quantified the mAP of object detection models trained on both manual and automated labeling methods, for three different sensor types. The discussion will now return to the two research questions identified earlier to examine key findings.

*How effective is MATT compared to manual labeling approaches?*

Given the speed of model training with the automated method versus the manual method, we find that MATT is highly effective at conducting image extraction, image processing, segmentation, and labeling. For example, using MATT reduces time requirements to labeling multispectral datasets by 87.8%, while only losing 6.7% in detection performance when compared to the labor intensive and time-consuming manual approaches. Furthermore, MATT can segment and label multispectral images with relatively high accuracy in .22 hours using a GPU and 1.1 hours with a CPU. If object detection applications require the highest precision necessary, then the manual method is best suited and will generally outperform the automated method by an average of 6.7%. However, the time requirements to manually label multispectral data necessitate 87.8% more time (20.0 hours for 2,400 images) when compared to time required to train a MATT model (2.4 hours for 2,400 images).

TABLE III
Time Requirements for Manual and MATT Dataset Labeling

| Task | Time (Hours) |
|---|---|
| MATT + CPU | 1.1 |
| MATT + GPU | 0.2 |
| MATT Manual Check | 2.2 |
| Manual Labeling | 20.0 |

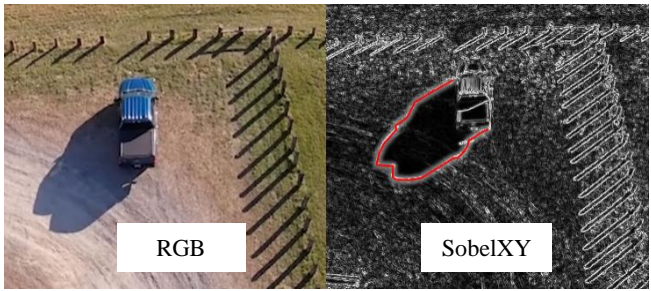

**Fig. 9.** Shadow from the RGB truck image appears in the SobelXY transformation (highlighted in red), causing edge distortion of the object class.

The time to manually label 2,400 images assumes that one individual is conducting labeling with no breaks taken during the process. Since humans require rest and typically work an average 8-hour day, manually labeling a dataset would realistically require significant more time. If one individual worked an 8-hour day and took one hour for lunch and one hour of combined breaks, this would leave 6 productive hours. With six productive hours in a day, it would take one human approximately 3.3 days to label a 2,400 image dataset. Thus, running MATT via a GPU saves 87.8% more time than creating a training dataset with the manual method. The benefits of using the automated method will increase exponentially as the number of images to segment and label also increases.

*What impact does elevation, time-of-day and sensor type have on models trained either via a manually labeled dataset or MATT?*

Elevation affects all the models differently. Although mAP performance loss is expected at higher elevations, it is counterintuitive that the lowest elevation of 16 m suffered from low detection performance. The primary reason for the low average mAP at 16 m was because of particularly poor LWIR performance in Pre-Sunrise and Post-Sunrise periods, which brought down the overall mAP of this elevation.

As previously mentioned, both the RGB manual and automated models have a gradual decrease in performance over elevation (which is to be expected). However, LWIR and RGB-LWIR models performed in a surprisingly dysfunctional manner at the lowest elevation of 16 m (21.4%). This may be because of early morning ground and object-class temperatures, which provide a lack of thermal contrast for strong edge detection. Furthermore, when analyzing previous research that used the same dataset, LWIR and RGB-LWIR performance was approximately 5% higher at 47 m during Pre-Sunrise and Post-Sunrise periods when compared to 16 m model performance. At Pre-Sunset and Post-Sunset periods, both the LWIR and RGB-LWIR models had consistently poor performance. Thus, there is a likely correlation between object class temperature, ground temperature, and elevation that affects LWIR and RGB-LWIR object detection models. The elevation that had the best overall model performance was 47 m (81.8% daytime, 69.9% nighttime). Indeed, 47 m therefore appears to be the optimal elevation that provides the best LWIR resolution for object detection due to good thermal contrast at this elevation.

The manual models outperformed automated models at night due to the difficulty of conducting segmentation during nighttime with RGB images. A primary limitation with the automated models is that the quality of LWIR and RGB-LWIR segmentation relies heavily on RGB segmentation. Thus, it is highly recommended to conduct multispectral data collection at noon. Noon will provide sufficient visibility and limited shadows for effective RGB segmentation, thereby providing accurate segmentation masks for LWIR and RGB-LWIR.

Within the daytime period, all models achieved peak performance during Noon time (67.5% mAP). This is most likely due to the minimal presence of shadows during Noon time images and footage. Shadows cast from object classes at Post-Sunrise and Pre-Sunset periods leads to a decrease in edge resolution. These shadows also created entirely new edges, which were inaccurate (as seen in the SobelXY image in Fig. 9). During model training, image augmentation is conducted within the neural network on the training dataset, such as SobelXY, to help the model identify and learn unique edges.

Diminishing edge resolution combined with newly produced incorrect edges formed by shadows led to an overall decrease in model mAP. Models trained with the automated method will be more susceptible to over-segmentation, or under segmentation, from data collected at Post-Sunrise or Pre-Sunset periods. Fig. 10 is an example of an image captured during Pre-Sunset period, resulting in longer shadows being captured and segmented by MATT as part of the object class. As a result, some post-processing had to be carried out to remove or add to the segmentation masks to ensure that the models would be trained on accurate ground-truth data. In Fig. 10, the red polygons are shadows that caused MATT to over segment the object class, while the orange polygon is a car that was under-segmented because of the shadow being cast on the hood, which prevented SAM from identifying it as part of the car. Applying RGB segmentation mask that derived from Post-Sunrise and Pre-Sunset periods to multispectral images would risk possibly applying inaccurate segmentation masks to train multispectral models.

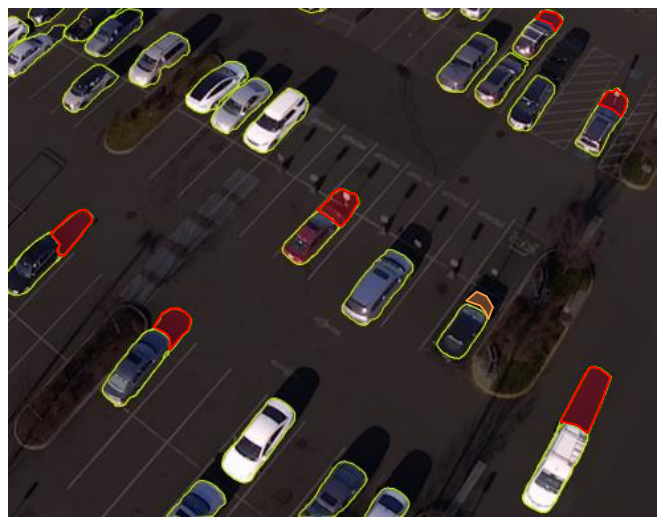

**Fig. 10.** The areas highlighted in red result from over-segmentation, while areas highlighted in orange are under-segmented, due to excessive shadows in images that are collected too close to Post-Sunrise or Pre-Sunset periods.

When analyzing nighttime average model performance, LWIR models performed perform better during Post-Sunset hours because object classes are sufficiently warmed due to thorough sun exposure throughout the day. In Pre-Sunrise periods, only certain portions of the object class would be warm, such as tires and the hood of the vehicle, leading to incomplete LWIR object class edges. During Post-Sunset periods, the body of the vehicle is generally fully outlined, thereby providing enhanced edge detection, leading to improved model performance. Although data was not collected at midnight, testing the models against LWIR data from midnight hours would provide an additional data point to further support the aforementioned theory.

Despite MATT models falling slightly behind manual model performance, MATT provides limitless automation potentials that will greatly enhance the field of multispectral object detection. With its automation capabilities MATT can create enormous multispectral datasets with minimal resources. MATT will be able to automate the most human-intensive task of machine learning model training, providing benefits from students embarking on multispectral research, all the way to governments, companies and organizations that rely heavily on multispectral applications.

## V.  CONCLUSION

This study set out to create a method that allows for the seamless transfer of segmentation masks from RGB images to co-aligning multispectral images, enabling efficient and time saving multispectral object detection model training. There are two key contributions to the literature, including (i) successfully creating a workflow that automatically segments and labels object classes in multispectral imagery, and (ii) quantifying how models trained with SAM perform using multispectral sensors from air-based platforms operating between 16-121 m.

MATT provides an easy-to-use solution to automatically label and segment multispectral datasets efficiently with significantly less human input, allowing for consistent labeling results at machine speed. For example, compared to a manual labeling approach for 2,400 images taking 20.0 hours, MATT can reduce this time commitment to 2.4 hours, with only an average precision loss of 6.7%. We believe this is sacrifice worth making to speed up the process of model experimentation and discovery using multispectral imagery. Indeed, the concept behind MATT can help researchers using multispectral imagery to rapidly prototype new ideas and concepts.

The methods developed in this study provide a myriad of opportunities for future multispectral research. For example, determining how both manual and automated labeling techniques perform when applied to different neural networks and object classes. Currently, the performance metrics from this research are based off a single pre-trained CNN, YOLOv8s. Lastly, the methods developed for MATT can be easily applied to other parts of the electromagnetic spectrum, such as Near-Infrared (NIR), Medium-Wave Infrared (MWIR), and HSI. Another area for future research is to combine sensors capable of adaptable luminosity with MATT. Adaptable luminosity is a sensor's ability to adjust fusion levels between different spectral bands to maximize resolutions given ambient illumination and temperature. Different spectral bands within the infrared spectrum, such as LWIR, can be fused at various concentrations to maximize object detection performance. For example, at zero percent illumination LWIR would not be fused with RGB. Conversely, at noon-time RGB would require minimal to no LWIR fusion. There is ample room for fusion experimentation between the periods of darkness and noon to determine optimal spectral fusion that enhances object detection performance.

Reflecting on the research, we identify two key limitations. Firstly, collection could not be conducted above 121 m because of current civilian airspace restrictions in the USA. Secondly, the approach adopted here focused only on measuring model performance for multispectral air-based imagery, providing various future research avenues for exploring hyperspectral and ground-based approaches. Certainly, as multispectral object detection continues to grow in coming years, there will be commensurate interest in speeding up current machine learning methods and techniques to build new models efficiently, increasing interest in the research reported here.

## ACKNOWLEDGMENT

This material is based upon work supported by the NSF National Center for Atmospheric Research, which is a major facility sponsored by the U.S. National Science Foundation under Cooperative Agreement No. 1852977. The project upon which this article is based was funded through the NSF NCAR Early-Career Faculty Innovator Program under the same Cooperative Agreement. EO would like to thank the Geography & Geoinformation Science Department at George

Mason University for providing start-up funding which purchased the LWIR sensor that enabled this research to take place.

DATA

To access the data and codebase used in this research, please use the hyperlinks below.

[GitHub](#)
[Data](#)

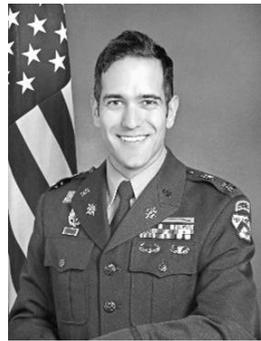

**JAMES E. GALLAGHER** received a B.A from Mercyhurst, an M.A from American Military University, and an MS from George Mason University. He is currently a PhD. student at George Mason studying multi-spectral object detection for drone-based applications. He is also an active-duty U.S. Army Major in the Military Intelligence Corps. MAJ Gallagher's research in RGB-LWIR object detection was inspired from his lessons-learned in Kirkuk Province, Iraq, where he used a variety of drone-based sensors to locate Islamic State insurgents.

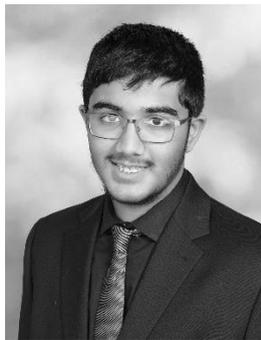

**ARYAV GOGIA** is a member of George Mason University's Aspiring Scientists Summer Internship Program (ASSIP). His research is primarily focused on automating machine learning tasks and building synthetic datasets for multispectral object detection models. Aryav also builds and prototypes autonomous data collection systems for environmental protection applications.

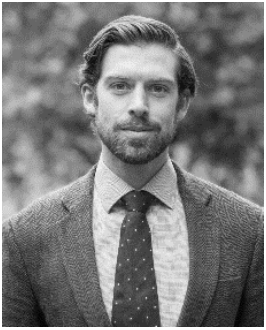**EDWARD J. OUGHTON** received the M.Phil. and Ph.D. degrees from Clare College, at the University of Cambridge, U.K., in 2010 and 2015, respectively. He later held research positions at both Cambridge and Oxford. He is currently an Assistant Professor in the College of Science at George Mason University, Fairfax, VA, USA, developing open-source research software to analyze digital infrastructure deployment strategies. He received the Pacific Telecommunication Council Young Scholars Award in 2019, Best Paper Award 2019 from the Society of Risk Analysis, and the TPRC48 Charles Benton Early Career Scholar Award 2021.